\documentclass[10pt,twocolumn,letterpaper]{article}

\usepackage[pagenumbers]{cvpr} 

\usepackage{graphicx}
\usepackage{amsmath}
\usepackage{amssymb}
\usepackage{booktabs}
\usepackage[accsupp]{axessibility}  

\usepackage{wrapfig}
\usepackage{tikz}
\usepackage{comment}
\usepackage{color}
\usepackage{graphicx}
\usepackage{multirow}
\usepackage{tabularx}
\usepackage{bbding}
\usepackage{bbm}
\usepackage{float}
\usepackage{cite}
\usepackage{xspace}
\usepackage[misc]{ifsym}
\usepackage{bbding}
\usepackage{pifont}
\usepackage{color,xcolor}
\usepackage{stfloats}
\usepackage{wasysym}
\newcommand{\cmark}{\ding{51}\xspace}%
\newcommand{\xmarkg}{\textcolor{lightgray}{\ding{55}}\xspace}%
\usepackage{algorithm,algpseudocode}

\newcounter{algsubstate}
\renewcommand{\thealgsubstate}{\alph{algsubstate}}
\newenvironment{algsubstates}
  {\setcounter{algsubstate}{0}%
   \renewcommand{\State}{%
     \stepcounter{algsubstate}%
     \Statex {\footnotesize\thealgsubstate:}\space}}
  {}

\newcommand{\ours}{\textbf{PADing}~\!(ours)\xspace}
\newcommand{\PAD}{\textbf{PADing}\xspace}

\usepackage{algorithm}  
\usepackage{algpseudocode}  
\usepackage{amsmath}

\usepackage[pagebackref,breaklinks,colorlinks]{hyperref}

\usepackage[capitalize]{cleveref}
\crefname{section}{Sec.}{Secs.}
\Crefname{section}{Section}{Sections}
\Crefname{table}{Table}{Tables}
\crefname{table}{Tab.}{Tabs.}

\def\cvprPaperID{2265} 
\def\confName{CVPR}
\def\confYear{2023}

\begin{document}

\title{Primitive Generation and Semantic-related Alignment\\for Universal Zero-Shot Segmentation}

\author{
Shuting He$^1$\footnotemark[2]
\qquad
Henghui Ding$^2$\footnotemark[2]~~$^{\textrm{\Letter}}$
\qquad
Wei Jiang$^1$\\
$^1$Zhejiang University
\qquad
$^2$Nanyang Technological University\\
\href{https://henghuiding.github.io/PADing}{https://henghuiding.github.io/PADing}
}
\maketitle
\renewcommand{\thefootnote}{\fnsymbol{footnote}}
\footnotetext[2]{Equal contribution.}
\footnotetext[0]{${\textrm{\Letter}}$ Corresponding author 
(henghui.ding@gmail.com).}

\begin{abstract}
We study universal zero-shot segmentation in this work to achieve panoptic, instance, and semantic segmentation for novel categories without any training samples. Such zero-shot segmentation ability relies on inter-class relationships in semantic space to transfer the visual knowledge learned from seen categories to unseen ones. Thus, it is desired to well bridge semantic-visual spaces and apply the semantic relationships to visual feature learning. We introduce a generative model to synthesize features for unseen categories, which links semantic and visual spaces as well as addresses the issue of lack of unseen training data. Furthermore, to mitigate the domain gap between semantic and visual spaces, firstly, we enhance the vanilla generator with learned primitives, each of which contains fine-grained attributes related to categories, and synthesize unseen features by selectively assembling these primitives. Secondly, we propose to disentangle the visual feature into the semantic-related part and the semantic-unrelated part that contains useful visual classification clues but is less relevant to semantic representation. The inter-class relationships of semantic-related visual features are then required to be aligned with those in semantic space, thereby transferring semantic knowledge to visual feature learning. The proposed approach achieves impressively state-of-the-art performance on zero-shot panoptic segmentation, instance segmentation, and semantic segmentation.
\end{abstract}

\section{Introduction}

\begin{figure}[t]
    \centering
	\includegraphics[width=0.49\textwidth]{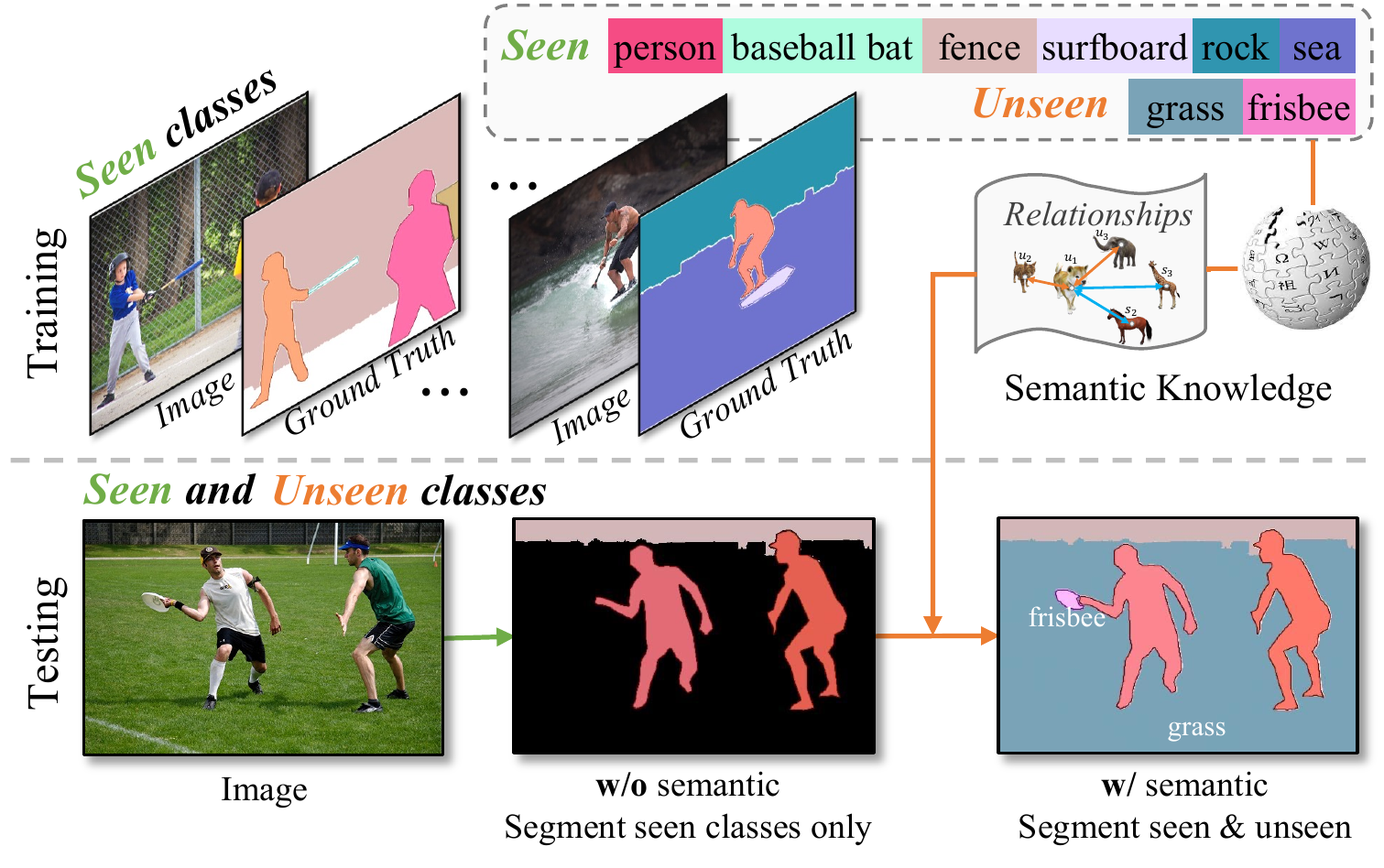}
	\vspace{-0.76cm}
	\caption{Zero-shot image segmentation aims to transfer the knowledge learned from seen classes to unseen ones (\ie, never shown up in training) with the help of semantic knowledge.} 
	\vspace{-0.16cm}
	\label{fig:teaser}
\end{figure}

Image segmentation aims to group pixels with different semantics, \eg, category or instance~\cite{TransformerSurvey,mask2former}. Deep learning methods~\cite{long2015fully,chen2017deeplab,maskrcnn,ding2020semantic,kirillov2019panoptic,PFPN,mask2former} have greatly advanced the performance of image segmentation with the powerful learning ability of CNNs~\cite{he2016deep} and Transformer~\cite{vaswani2017attention}. However, since deep learning methods are data-driven, great challenges are induced by the intense demand for large-scale labeled training samples, which are labor-intensive and time-consuming. To address this issue, zero-shot learning (ZSL)~\cite{lampert2009learning,palatucci2009zero} is proposed to classify novel objects with no training samples.
Recently, ZSL is extended into segmentation tasks like zero-shot semantic segmentation (ZSS)~\cite{ZS3,SPNet} and zero-shot instance segmentation (ZSI)~\cite{zsi}.
Herein, we further introduce zero-shot panoptic segmentation (ZSP) and aim to build a universal framework for zero-shot panoptic/semantic/instance segmentation with the help of semantic knowledge, as shown in \cref{fig:teaser}.

Different from image classification, segmentation requires pixel-wise classification and is more challenging in terms of class representation learning. Substantial efforts have been devoted to zero-shot semantic segmentation~\cite{SPNet,ZS3}
and can be categorized into projection-based methods~\cite{SPNet,ding_iccv21,ZegFormer} and generative model-based methods~\cite{ZS3,CSRL,CaGNet}.
The generative model-based methods are usually superior to the projection-based methods because they produce synthetic training features for the unseen group, which contribute to alleviating the crucial bias issue~\cite{reviewGZSL} of tending to classify objects into seen classes. Owing to the above merits, we follow the paradigm of generative model-based methods to address zero-shot segmentation tasks. 

However, the current generative model-based methods are usually in the form of per-pixel-level generation, which is not robust enough in the more complicated scenarios. Recently, several works propose to decouple the segmentation into class-agnostic mask prediction and object-level classification \cite{mask2former,solov1,chen2022weak,D2Zero}. We follow this strategy and degenerate the pixel-level generation to a more robust object-level generation.
What's more, previous generative works~\cite{ZS3,CSRL,CaGNet} usually learn a direct mapping from semantic embedding to visual features. Such a generator does not consider the visual-semantic gap of feature granularity that images contain much richer information than languages. The direct mapping from coarse to fine-grained information results in low-quality synthetic features. To address this issue, we propose to utilize abundant primitives with very fine-grained semantic attributes to compose visual representations. Different assemblies of these primitives construct different class representations, where the assembly is decided by the relevance between primitives and semantic embeddings.
Primitives greatly enhance the expressive diversity and effectiveness of the generator, especially in terms of rich fine-grained attributes, making the synthetic features for different classes more reliable and discriminative.

However, there are only real image features of seen classes to supervise the generator, leaving unseen classes unsupervised.
To provide more constraints for the feature generation of unseen classes, we propose to transfer the inter-class relationships in semantic space to visual space. 
The category relationships obtained by semantic embeddings are employed to constrain the inter-class relationships of visual features. With such constraint, the visual features, especially the synthesized features for unseen classes, are promoted to have a homogeneous inter-class structure as in semantic space. Nevertheless, there is a discrepancy between the visual space and the semantic space~\cite{tong2019hierarchical,chen2021semantics}, so as to their inter-class relationships. Visual features contain richer information and cannot be fully aligned with semantic embeddings. Directly aligning two disjoint relationships inevitably compromises the discriminative of visual features. 
To address this issue, we propose to disentangle visual features into semantic-related and semantic-unrelated features, where the former is better aligned with the semantic embedding while the latter is noisy to semantic space. We only use semantic-related features for relationship alignment. The proposed relationship alignment and feature disentanglement are mutually beneficial. 
Feature disentanglement builds semantic-related visual space to facilitate relationship alignment and excludes semantic-unrelated features that are noisy for alignment. Relationship alignment in turn contributes to disentangling semantic-related features by providing semantic clues. 

Overall, the main contributions are as follows:
\vspace{-2mm}
\begin{itemize}
\setlength\itemsep{0em}
\item We study universal zero-shot segmentation and propose \textbf{P}rimitive generation with collaborative relationship \textbf{A}lignment and feature \textbf{D}isentanglement learn\textbf{ing} (\textbf{\PAD}) as a unified framework for ZSP/ZSI/ZSS.
\item We propose a primitive generator that employs lots of learned primitives with fine-grained attributes to synthesize visual features for unseen categories, which helps to address the bias issue and domain gap issue.

\item We propose a collaborative relationship alignment and feature disentanglement learning approach to facilitate the generator producing better synthetic features.
 
\item The proposed approach \textbf{\PAD} achieves new state-of-the-art performance on zero-shot panoptic segmentation (ZSP), zero-shot instance segmentation (ZSI), and zero-shot semantic segmentation (ZSS).
\end{itemize}

\section{Related Work}
\textbf{Zero-shot learning (ZSL)}~\cite{lampert2009learning,palatucci2009zero,kankuekul2012online,zhang2016zero} aims to classify images of unseen classes with no training samples via utilizing semantic descriptors as auxiliary information. There are two main paradigms: classifier-based methods that learn a visual-semantic projection~\cite{akata2015evaluation,li2018deep,zhang2016zero} and instance-based methods~\cite{dinu2014improving,yu2013designing} that synthesize fake samples for unseen classes. 
\textbf{Generalized zero-shot learning (GZSL)}, introduced by Scheirer et al.~\cite{scheirer2012toward}, aims to classify samples from both seen and unseen sets. Then, Chao et al.~\cite{chao2016empirical} show that the ZSL methods can't work well in GZSL setting from experiments, due to the feature of overfitting on seen classes. Classification score calibration methods~\cite{changpinyo2020classifier,das2019zero,guo2019dual,huynh2020fine} and out-of-distribution detector methods~\cite{bhattacharjee2019autoencoder, felix2019generalised} are proposed to alleviate this bias issue. 

\begin{figure*}[t]
    \centering
	\includegraphics[width=1.0\textwidth]{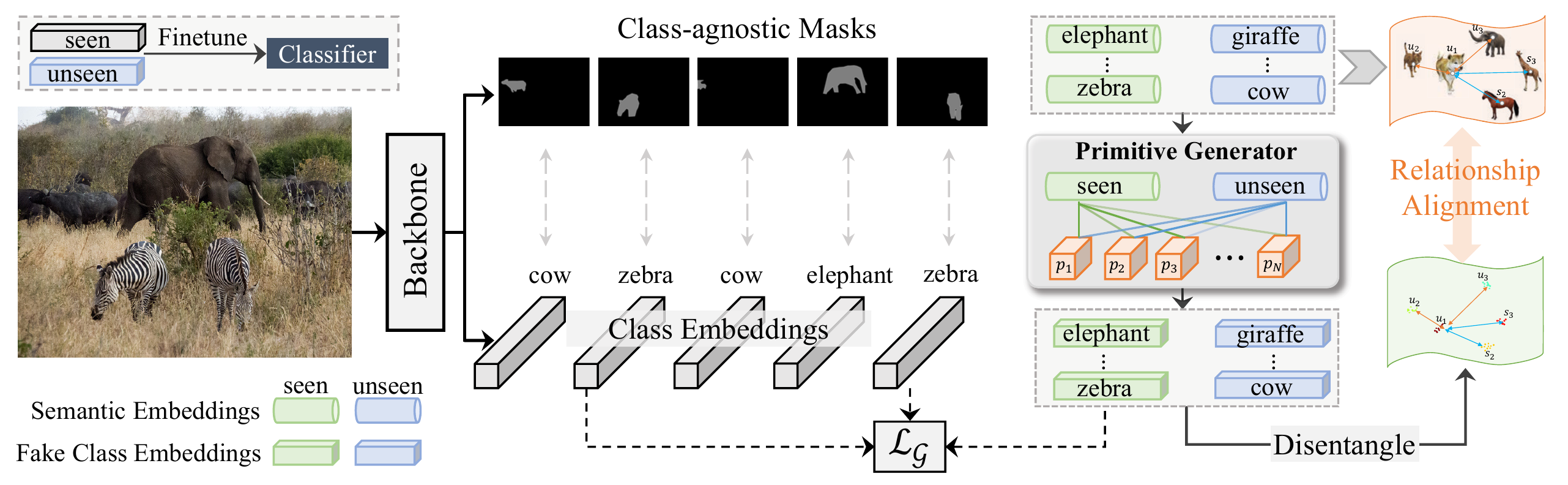}
	\vspace{-0.56cm}
    \caption{Overview of our approach \textbf{\PAD} for universal zero-shot image segmentation. We first obtain class-agnostic masks and their corresponding global representations, named class embeddings, from our backbone. A primitive generator is trained to produce synthetic features (\ie, fake class embeddings). The classifier, which takes class embeddings as input, is trained with both the real class embeddings from image and synthetic class embeddings by the generator. During the training of the generator, the proposed feature disentanglement and relationship alignment are employed to constrain the synthesized features.} 
	\vspace{-0.16cm}
	\label{fig:framework}
\end{figure*}

\textbf{Image Segmentation} is one of the most fundamental computer vision tasks~\cite{GRES,ding2019boundary,dong2023federated,ding2021vision,ding2022vlt,liu2022instance}. Deep-learning-based image segmentation methods under a fully supervised manner are extensively studied \cite{long2015fully,chen2017deeplab,liu2022few,maskrcnn,solov1,shuai2018toward,kirillov2019panoptic,PFPN,mask2former,ding2018context,ding2019semantic}. However, these methods require a large number of labeled training samples and cannot handle unseen categories that do not appear or are not defined in training data. To address these issues, \textbf{Zero-Shot Semantic Segmentation (ZSS)}~\cite{ZS3} and \textbf{Zero-Shot Instance Segmentation (ZSI)}~\cite{zsi} extend ZSL methods to semantic segmentation and instance segmentation, respectively. In this work, we further introduce \textbf{Zero-Shot Panoptic Segmentation (ZSP)} to extend the zero-shot learning to the panoptic segmentation task. There are two main paradigms: projection-based methods \cite{pastore2021closer,SPNet,ZegFormer,zsseg_baseline, lv2020learning, ding_iccv21, hu2020uncertainty, chen2022weak} and generative-based methods \cite{CSRL,ZS3,SIGN}. 
Projection-based techniques commonly utilize a projection approach to map the visual or semantic features of seen categories onto a shared space. (\eg, visual, semantic, or latent space), and then classify novel objects by measuring the feature similarity in the common space. The generative methods adopt generator to produce synthetic features for unseen classes. However, existing generative works~\cite{ZS3,CSRL,CaGNet} usually learn a direct mapping from semantic embedding to visual features and do not consider the visual-semantic gap of feature granularity. We design a primitive generation and semantic-related alignment approach to universally address zero-shot segmentation, including ZSP, ZSI, and ZSS.

\section{Methodology}
\cref{fig:framework} illustrates the overview architecture of our proposed approach, \textbf{P}rimitive generation with collaborative relationship \textbf{A}lignment and feature \textbf{D}isentanglement learn\textbf{ing} (\textbf{\PAD}). Our backbone predicts a set of class-agnostic masks and their corresponding class embeddings. Primitive generator is trained to synthesize class embeddings from semantic embeddings. The real \& synthetic class embeddings are disentangled to semantic-related and semantic-unrelated features. We conduct the relationship alignment learning on the semantic-related feature. With the synthesized unseen class embeddings, we re-train our classifier with both the real class embedding of seen categories and the synthetic class embedding of unseen categories. The training process is demonstrated in \cref{alg:trainingUZSS}. The details of each part will be introduced in the following sections.

\subsection{Task Formulation}\vspace{-0.16cm}
Herein we give the problem formulation of zero-shot image segmentation. There are two spaces, feature space $\mathcal{X}$ and semantic space $\mathcal{A}$, to represent the visual features of images and semantic representations of categories, denoted as $\mathcal{X}\!=\!\{X^s, X^u\}$, $\mathcal{A}\!=\!\{A^s, A^u\}$, respectively. The superscript $s$ and $u$ represent the two non-overlapping groups, $N^s$ seen categories and $N^u$ unseen categories, respectively. We use $\mathcal{Y}\!=\!\{Y^s, Y^u\}$ to denote the ground truth label. $Y^s$ is label set of seen group and $Y^u$ is label set of unseen group, $Y^s\cap Y^u=\varnothing$. The training set is constructed from the images that contain any of the $N^s$ seen categories but no unseen categories, which is different from the open-vocabulary paradigm~\cite{OVRCNN,huynh2022open}.
According to the categories that appear in the testing set, there are two different settings named zero-shot learning (ZSL) and generalized zero-shot learning (GZSL). ZSL only classifies testing samples of unseen categories while GZSL needs to classify testing data of both seen and unseen categories. Zero-shot segmentation is naturally a kind of GZSL since the because images typically contain multiple and diverse categories. In this work, all the zero-shot segmentation tasks are under the GZSL setting unless otherwise specified.

\subsection{Primitive Cross-Modal Generation}

Due to the lack of unseen samples, the classifier cannot be optimized with features of unseen classes. As a result, the classifier trained on seen classes tends to assign all objects/stuff a label of seen group, which is called bias issue~\cite{chao2016empirical}. To address this issue, previous methods~\cite{ZS3,CSRL,CaGNet} propose to utilize a generative model to synthesize fake visual features for unseen classes. However, 
previous generative zero-shot segmentation works~\cite{ZS3,CSRL,CaGNet} commonly adopt Generative Moment Matching Network (GMMN) \cite{li2015generative,CSRL} or GAN~\cite{goodfellow2020generative}, which consist of multiple linear layers as feature generator. Such a generator, though achieves good performance, does not consider the visual-semantic difference of feature granularity. It is well known that image generally contains much richer information than language. Visual information provides very fine-grained attributes of objects while textual information typically provides abstract and high-level attributes. Such difference results in an inconsistency between visual features and semantic features. To address this challenge, we propose a Primitive Cross-Modal Generator that employs lots of learned attribute primitives to construct visual representations.

\begin{figure}[t]
    \centering
	\includegraphics[width=0.436\textwidth]{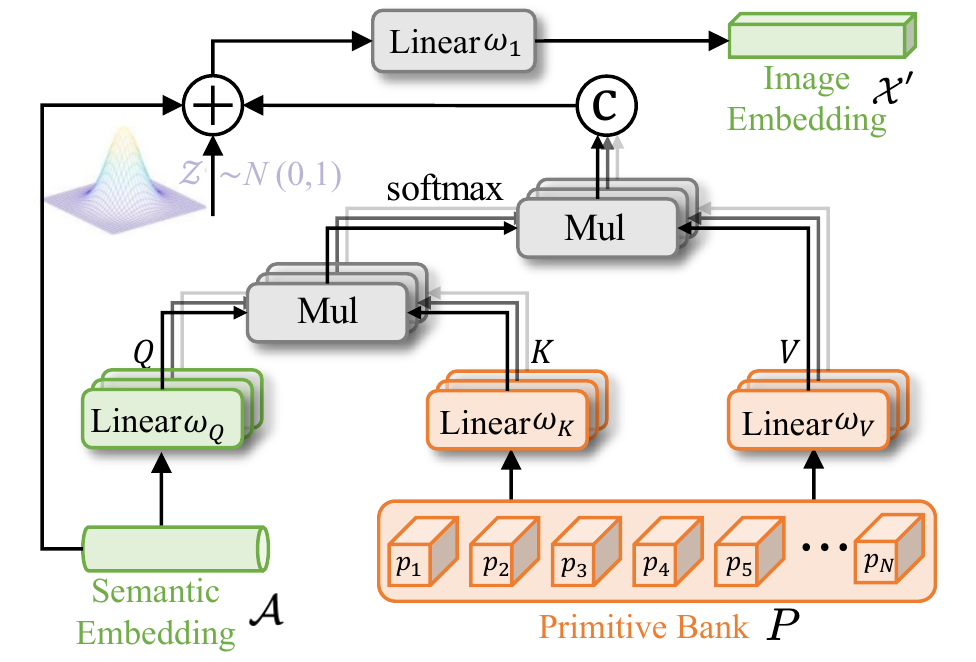}
	\vspace{-0.16cm}
	\caption{Primitive Cross-Modal Generator. We use lots of learned primitives to represent fine-grained attributes. The generator synthesizes visual features via assembling these primitives according to input semantic embedding.}
	\label{fig:generator}
\end{figure}

As shown in \cref{fig:generator}, we build our Primitive Generator with a Transformer architecture. First, a set of learnable primitives are randomly initialized, denoted as $P=\{p_i\}_{i=1}^N$, where $p_i\in\mathbb{R}^{d_k}$ and $d_k$ is the number of channels. These primitives are assumed to contain very fine-grained attributes related to categories, \eg, hair, color, shape, \etc. Different kinds of assembly of these primitives build different representations for categories. A self-attention is first performed on these primitives to construct relationship graph among these primitives. Next, we utilize two different linear layers $\omega_K$ and $\omega_V$ to deal with $P$ to obtain the Key and Value for cross attention, denoted as $K$ and $V$ respectively. Then, taking semantic embeddings as Query $Q$, cross attention is performed as 
\begin{equation}\label{eq:mha}
\small
   {\mathcal{X}}' =  \omega_1\left(\mathrm{softmax}(\frac{\mathcal{Q}K^{T}}{\sqrt{d_k}})V+\mathcal{A}+\mathcal{Z}\right),
\end{equation}
where ${\mathcal{X}}'$ represents synthetic visual features and $\mathcal{Z}$ denotes random sample with a fixed Gaussian distribution. $\omega_1$ is the linear layer. Different from feature generation via processing semantic embedding with several linear layers, we synthesize visual features via weighted assembling these abundant primitives, which provides much more diverse and richer representations. Moreover, for related categories that share some similarities in semantic space, primitives provide an explicit way to express such similarities. For example, \texttt{dog} and \texttt{cat} both have the attributes of hairy and tail, so the primitives related to hairy and tail show high response to the semantic embedding query of \texttt{dog} and \texttt{cat}. With such primitives that describe fine-grained attributes, we can easily construct different category representations and transfer the knowledge of seen classes to unseen ones.

We follow~\cite{li2015generative} to define our generator loss $\mathcal{L}_\mathcal{G}$ to diminish maximum mean discrepancy between two probability distributions:
\begin{equation}
\footnotesize
\mathcal{L}_\mathcal{G}=\sum_{f,\dot{f}\in X^s} k(f, \dot{f}) + \sum_{f',\dot{f}'\in {X^s}'} k(f',\dot{f}') - 2\sum_{f \in X^s} \sum_{f'\in {X^s}'} k(f,f'),
\label{Eq:losslg}
\end{equation}
where $X^s$ and ${X^s}'$ denote real visual features and synthetic visual features of seen classes, respectively. $k$ is a kernel and $k(f, f') = \exp(-\frac{1}{2\sigma^2} \|f-f' \|^{2})$ with bandwidth $\sigma$.

When a semantic embedding from unseen group is fed into the trained Primitive Generator, we can get its corresponding synthetic class embedding. We then re-train our classifier with both the real class embedding of seen categories and the synthetic class embedding of unseen categories, which greatly alleviates the bias issue.
Besides, such global representations are more robust than per-pixel classification \cite{zsi,ZS3,ding_iccv21,CaGNet,CSRL} and can thus have a better alignment between visual space and semantic space.

\begin{algorithm}[t]
 \footnotesize
  \caption{\small Training of Universal Zero-Shot Segmentation.}
  \label{alg:trainingUZSS}  
  \begin{algorithmic}[1]  
    \Require Image $I^s$ of seen classes, semantic embedding $\mathcal{A}=\{A^s,A^u\}$ from text encoder; 
    \State Pre-train the visual backbone with images of seen classes;
    \State Forward image $I^s$ into the trained visual backbone, generating a set of masks $M^s$ and their corresponding class embeddings $X^s$;
    \State Train primitive generator:
        \begin{algsubstates}
            \State Forward semantic embedding $\mathcal{A}$ into primitive generator to get synthetic visual feature $\mathcal{X}'=\{{X^s}',{X^u}'\}$, supervise ${X^s}'$ with $X^s$ by $\mathcal{L}_\mathcal{G}$ in \cref{Eq:losslg};

            \State Disentangle visual feature $x_i \in \{X^s~\mathrm{or}~\mathcal{X}'\}$ into semantic-related feature $\hat{x}_i$ and semantic-unrelated feature $\ddot{x}_i$;
            
            \State Constrain semantic-related feature $\hat{x}_i$ using alignment $\mathcal{L}_{\mathbb{A}}$;
        \end{algsubstates}
    \State Fine-tune the classifier of the visual backbone with $X^s$ and ${X^u}'$.
  \end{algorithmic}
\end{algorithm}
\subsection{Semantic-Visual Relationship Alignment}

\begin{figure}[t]
    \centering
	\includegraphics[width=0.46\textwidth]{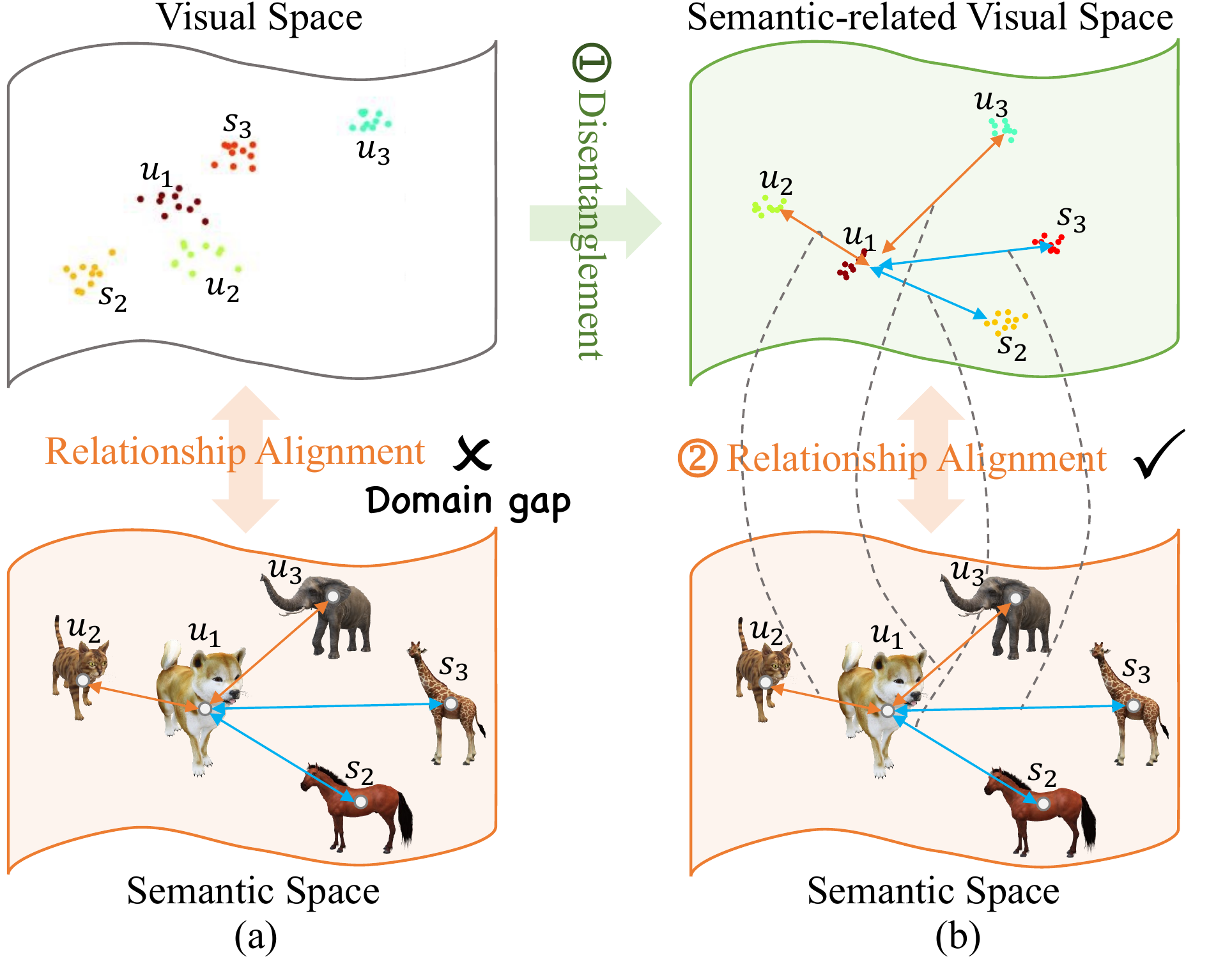}
	\vspace{-0.26cm}
	\caption{\textbf{Relationship alignment}. (a) The conventional relation-ship alignment. (b) Our proposed two-step relationship alignment. Considering the domain gap, we introduce a semantic-related visual space, where features are disentangled from visual space and have more direct relevance with semantic space. We have the relationship in semantic-related visual space be aligned with semantic space. $u_i$/$s_j$ refers to unseen/seen category. Taking $u_1$ \texttt{dog} as an example, we aim to transfer its similarities with \{\texttt{cat}, \texttt{elephant}, \texttt{horse}, \texttt{zebra}\} from semantic to visual space. } 
	\vspace{-0.16cm}
	\label{fig:relationship}
\end{figure}

It is well known that relationships among categories are naturally different \cite{CSRL, vyas2020leveraging,chen2021weak}. For example, there are three objects: \texttt{apple}, \texttt{orange}, and \texttt{cow}. Obviously, the relationship of \texttt{apple} \& \texttt{orange} is closer than \texttt{apple} \& \texttt{cow}. Class relationships in semantic space are powerful prior knowledge, while the category-specific feature generation does not explicitly leverage such relationships. As shown in \cref{fig:relationship}, we build such relationships with semantic embeddings and explore to transfer this knowledge to visual space, making semantic-visual alignment in terms of class-wise relationships. By considering the relationship, there are more constraints on the unseen categories' feature generations, to pull or push their distances with seen categories.

\vspace{-0.3cm}
\paragraph{Semantic-related Visual Feature} However, the visual features are not fully aligned with the semantic representations but contain richer information including semantic-related visual features and also semantic-unrelated visual features. Semantic-unrelated features may have strong visual clues and contribute to classification, but have low relevance with language semantic representations. Directly aligning semantic embeddings with original visual features would confuse the generator and reduce its generalization to unseen categories. To address this issue, we propose to disentangle the semantic-related visual features and semantic-unrelated visual features. 
Given a feature $x_i$, where $x_i\in \mathcal{X}$ is the class embedding  from either backbone or our generator, feature disentanglement learns how to disentangle and reconstruct $x_i$ itself.
We use encoder $E_{\mathbb{R}}$ to extract semantic-related feature, $\hat{x}_i=E_{\mathbb{R}}(x_i)$.  
Then, we calculate the correlation score between semantic-related feature $\hat{x}_{i}$ and semantic embeddings $\mathcal{A}=\{a_1,...,a_{N^s+N^u}\}$.
$E_{\mathbb{R}}$ is trained with cross-entropy loss as a classification problem to endow semantic-related features $\hat{x}_{i}$ with discriminative semantic knowledge, \ie, 
\begin{equation}
\label{eq:lr}
\small
    \mathcal{L}_\mathbb{R} = -\sum_{i}\sum_{k}\mathbbm{1}([\hat{x}_i]=k)log\frac{\mathrm{exp}(\hat{x}_{i}a_k/\tau)}{\sum_{k}\mathrm{exp}(\hat{x}_{i}a_k/\tau)},
\end{equation}
where $[\hat{x}_i]$ is the ground truth class intex of $\hat{x}_i$, $\mathbbm{1}(\cdot)$ is the indicator function that outputs 1 if the condition is true and 0 otherwise. $\tau$ is the temperature parameter.

We use another encoder $E_{\mathbb{U}}$ to extract semantic-unrelated feature, denoted as $\ddot{x}_i=E_{\mathbb{U}}(x_i)$. We suppose the semantic-unrelated features to have the normal distribution $\mathcal{N}(0,1)$ with zero mean and unit variance \cite{lee2018diverse}. We use KL divergence loss to constrain the distribution range,
\begin{equation}
 \small
\mathcal{L}_\mathbb{U}= \sum_{i} D_{KL}[\ddot{x}_i||\mathcal{N}(0,1)],
\label{eq:eq3}
\end{equation}
where $D_{KL}[p||q]$~\!$=$~\!$-\int p(z) log\frac{p(z)}{q(z)}$. Such that each class has its own independent and diverse feature component.
To push the network to extract more representative semantic-related features and preserve visual feature information, we reconstruct the feature with a decoder $D$ under $\ell_1$ loss: 
\begin{equation}
 \small
    \mathcal{L}_{recon}  = \sum_{i}\left\lVert x_i - D(\hat{x}_{i}, \ddot{x}_{i}) \right\rVert_1.
\end{equation}
The training objective for feature disentanglement is $\mathcal{L}_\mathbb{D}=\mathcal{L}_\mathbb{R}+\mathcal{L}_\mathbb{U}+\mathcal{L}_{\mathrm{recon}}$.

\vspace{-0.3cm}
\paragraph{Relationship Alignment} Then we conduct relationship alignment between semantic-related visual space and semantic space. We use KL divergence loss to make the similarity of any two semantic-related features $\hat{x}_i$ and $\hat{x}_j$ reach the similarity of their corresponding semantic embeddings $a_{[\hat{x}_i]}$ and $a_{[\hat{x}_j]}$, \ie,
\begin{equation}
\small
  \mathcal{L}_\mathbb{A} =  D_{KL}[\frac{\hat{x}_i\hat{x}_j}{\lVert\hat{x}_i\rVert\lVert\hat{x}_j\rVert}/\tau||\frac{a_{[\hat{x}_i]}a_{[\hat{x}_j]}}{\lVert a_{[\hat{x}_i]}\rVert\lVert a_{[\hat{x}_j]}\rVert}/\tau],
\label{eq:eq6}
\end{equation}
where $[\hat{x}_i]$ is the ground truth class index of $\hat{x}_i$, $\tau$ is the temperature parameter to control the sharpness of similarity distribution operating on the KL loss. $\hat{x}_i^s$ of the seen group is from either real features or synthetic features while $\hat{x}_i^u$ of the unseen group is from synthetic features by generator only. There are two kinds of alignment, intra-group alignment and inter-group alignment, with different focuses in \cref{eq:eq6}.
When $\hat{x}_i$ and $\hat{x}_j$ are from the same group, \eg, $\hat{x}_i^s$ and $\hat{x}_j^s$ both from seen group, it is intra-group alignment and contributes to extracting better class representations with the relationships as a constraint. When they are from different groups, \eg, $\hat{x}_i^s$ from seen group and $\hat{x}_j^u$ from unseen group, it is inter-group alignment that aims to transfer the relationship knowledge from seen to unseen. Inter-group alignment gives constraints on the relationships of seen and unseen categories, real features and synthetic features. It greatly improves the model's adaptability and generalization to unseen categories. 

\vspace{2mm}
\noindent\textbf{Collaborative Disentanglement and Alignment} 
Our disentanglement and alignment are complementary and mutually beneficial. On the one hand, disentanglement promotes relationship alignment. With the disentanglement, semantic-related features can be extracted for alignment and semantic-unrelated noises are excluded. 
On the other hand, relationship alignment facilitates disentanglement. Introducing intra-group and inter-group alignment, class-wise relationship among semantic-related features can be constructed and the discrepancy between semantic-visual feature distributions can be reduced, eventually leading to the improvement of the feature disentanglement.

\subsection{Training Objective}
\Cref{alg:trainingUZSS} shows the overall training pipeline of our universal zero-shot segmentation model. First, we pre-train our segmentation backbone with annotated data from seen classes in a full-supervision manner.
Next, We train the primitive generator under the following objective:
\begin{equation}
 \mathcal{L}_{total} =  \mathcal{L}_\mathcal{G} + \lambda (\mathcal{L}_\mathbb{D}+ \mathcal{L}_\mathbb{A}),
\label{eq:final_loss}
\end{equation}
where $\lambda$ is the weight to control the importance of the disentanglement and alignment module.
Once the generator is trained above, it can generate synthetic features for unseen classes. Together with the real features from seen classes, we can train a new classification layer.

\section{Experiments}
\subsection{ Experimental Setup}

\textbf{Implementation Details.} The proposed network and all our experiments are implemented based on Pytorch. We utilize CLIP text embeddings~\cite{CLIP} and word2vec~\cite{mikolov2013distributed} as our semantic embedding and normalize it with $\ell_2$ normalization. 
CLIP text embeddings are extracted following the previous works~\cite{ZegFormer,vild}.
We adopt Mask2Former \cite{mask2former} build upon the ResNet-50 as backbone~\cite{he2016deep}, with 100 queries for both training and inference. Hyper-parameters are consistent with the setting of \cite{mask2former} unless otherwise specified. Encoder 
$E_{\mathbb{R}}$ and $E_{\mathbb{U}}$ are both multi-layer perceptron (MLP) containing one hidden layer, LeakyReLU activation and dropout. $E_{\mathbb{D}}$ is constructed with two stacked single MLP layers followed by LeakyReLU activation and dropout.
We apply SGD optimizer for the parameters of classifier with learning rate $1\times10^{-3}$, weight decay $5\times10^{-4}$ and momentum 0.9, and Adam optimizer for the parameters of generator, $E_{\mathbb{R}}$, $E_{\mathbb{U}}$, and $E_{\mathbb{D}}$ with initial learning rate $2\times10^{-4}$.
The number of the Transformer layers, loss weight $\lambda$ in \cref{eq:final_loss}, temperature $\tau$, $\sigma$ are set to 3, 0.002, 0.1, \{2, 5, 10, 20, 40, 60\}, respectively.

\textbf{Datasets.} We use the popular dataset MSCOCO 2017, which consists training set with 118k images and validation set with 5k images. For panoptic segmentation, 133 classes (80 \texttt{thing} classes and 53 \texttt{stuff} classes) are included in annotations. 
For semantic segmentation, COCO-Stuff contains 171 valid classes in total. To get a fair comparison with ZSI~\cite{zsi}, we use MSCOCO 2014 for instance segmentation which contains 80k training and 40k validation images.

\subsection{Zero-Shot Panoptic Segmentation Task}
Because of the high similarities between semantic segmentation and panoptic segmentation, we develop the ZSP datasets by following the previous ZSS works~\cite{SPNet}. 
In order to avoid any information leakage, SPNet selects 15 classes in COCO stuff that do not appear in ImageNet as unseen classes. In COCO panoptic dataset, we find 14 classes overlapped with the 15 ones selected by SPNet and set them as unseen classes, \ie, \{\small\texttt{cow}, \texttt{giraffe}, \texttt{suitcase}, \texttt{frisbee}, \texttt{skateboard}, \texttt{carrot}, \texttt{scissors}, \texttt{cardboard}, \texttt{sky-other-merged}, \texttt{grass-merged}, \texttt{playingfield}, \texttt{river}, \texttt{road}, \texttt{tree-merged}\}\normalsize, while the remaining 119 classes are set as seen classes.
To guarantee no information leakage in the training set, we discard the training images that contain \textbf{even one pixel of any unseen classes}. Thus the model is trained by samples of seen classes only with 45617 training images. We use all 5k validation images to evaluate the performance of ZSP.
Panoptic and semantic segmentation tasks are evaluated on the union of \texttt{thing} and
\texttt{stuff} classes while instance segmentation is only evaluated on the \texttt{thing} classes.

\textbf{Evaluation Metrics.}
Under the GZSL setting, the model needs to segment objects/stuff of both seen and unseen classes, which is closer to real-world complicated scenarios. 
Following previous ZSS~\cite{SPNet,ZS3}, ZSD~\cite{bansal2018zero}, and ZSI~\cite{zsi} tasks, we compute seen metrics, unseen metrics, and the harmonic mean (HM) of seen metrics and unseen metrics as follows,
\vspace{-2mm}\begin{equation}\vspace{-1mm}\small
    \rm{HM} = \frac{2\times \rm{P}_{seen}\times \rm{P}_{unseen}}{\rm{P}_{seen}+\rm{P}_{unseen}},
\end{equation}
where $\rm{P}_{seen}$ and $\rm{P}_{unseen}$ denote the seen and unseen metrics, respectively. 
We use the PQ (panoptic quality) metric~\cite{kirillov2019panoptic} which can be viewed as the multiplication of a segmentation quality (SQ)  and a recognition quality (RQ).
We also report the results on instance segmentation, object detection and semantic segmentation tasks. For instance segmentation and object detection, we use the standard mAP (mean Average Precision)~\cite{ms_coco} with an IoU threshold of 0.5. For semantic segmentation, we use mIoU (mean Intersection-over-Union)~\cite{everingham2015pascal}.

\renewcommand{\multirowsetup}{\centering}
\begin{table*}[ht]
\setlength\tabcolsep{7.8pt}
\footnotesize
  \begin{center}
  \begin{tabular}{l|ccc|ccc|ccc|ccc}
  \hline
 &  & & & \multicolumn{3}{c|}{Seen} & \multicolumn{3}{c|}{Unseen} & \multicolumn{3}{c}{HM}\\

 \ \ \ \ Method &  G/P & A & D  &  PQ  & SQ & RQ  & PQ  & SQ & RQ &  PQ  & SQ & RQ \\
  \hline
  \hline
1) \textcolor{gray}{Supervised} & \xmarkg& \xmarkg&\xmarkg&\textcolor{gray}{43.2} & \textcolor{gray}{80.8} & \textcolor{gray}{51.6} &\ \ \textcolor{gray}{0.0} & \ \ \textcolor{gray}{0.0} &\ \ \textcolor{gray}{0.0} & \textcolor{gray}{-}& \textcolor{gray}{-} & \textcolor{gray}{-} \\
2) Projection & \xmarkg& \xmarkg&\xmarkg& 43.3 & 80.9 & 51.7& \ \ 0.0 &\ \ 0.0 & \ \ 0.0&\ \ 0.0 &\ \ 0.0 &\ \ 0.0  \\
3) Baseline &G& \xmarkg&\xmarkg&40.1 & 80.1 & 48.2 &\ \ 4.9 & 48.5 &\ \ 5.8 &\ \ 8.7 &60.4 & 10.3 \\
4) P-only & P& \xmarkg&\xmarkg&38.9 &79.9 & 46.2 & 11.5 & 56.5& 13.8& 17.7& 66.1&21.2\\
5) P\&A & P&\cmark& \xmarkg& 38.4 &79.2 & 45.8 & 13.8 & 52.7& 16.4& 20.3& 63.2 &24.1 \\
6) \textbf{\PAD} &P &\cmark&\cmark& \textbf{41.5} & \textbf{80.6} & \textbf{49.7} & \textbf{15.3} & \textbf{72.8} & \textbf{18.4} &\textbf{22.3} & \textbf{76.5} & \textbf{26.8}  \\
  \hline
  \end{tabular}
\vspace{-0.3cm}
\caption{Zero-shot panoptic segmentation ablation study results on MSCOCO. G, P, A, D denote GMMN generator, primitive generator, disentanglement, and alignment, respectively.}\label{tab:SOTA_1}
\vspace{-0.5cm}
\end{center}
\end{table*}
\renewcommand{\multirowsetup}{\centering}
\begin{table*}[t]
\setlength\tabcolsep{9.6pt}
\footnotesize
  \begin{center}
  \begin{tabular}{l|ccc|ccc|ccc|ccc}
  \hline
 &  &&  & \multicolumn{3}{c|}{Object Detection (ZSD)} & \multicolumn{3}{c|}{Instance Segmentation (ZSI)} & \multicolumn{3}{c}{Semantic Segmentation (ZSS)}\\

\ \ \ \ Method &  G/P & A & D &   Seen  & Unseen & HM  & Seen  & Unseen & HM &  Seen  & Unseen & HM  \\
  \hline
  \hline

1) \textcolor{gray}{Supervised} &\xmarkg&\xmarkg&  \xmarkg & \textcolor{gray}{53.2}&\ \ \textcolor{gray}{0.0} & \textcolor{gray}{-} & \textcolor{gray}{53.9}&\ \ \textcolor{gray}{0.0} &\textcolor{gray}{-} & \textcolor{gray}{51.2}  &\ \ \textcolor{gray}{0.0} &\textcolor{gray}{-} \\
2) Projection &\xmarkg&\xmarkg&  \xmarkg &53.2 & 12.6& 20.4 & 54.0& 12.5 & 20.3 & 50.8  &\ \ 1.2 &\ \ 2.3 \\
3) Baseline& G &\xmarkg& \xmarkg&52.2 & 16.5&25.0 & 52.8& 16.2&24.7 & 50.8  & 11.6 & 18.8\\
4) P-only&P& \xmarkg & \xmarkg & 52.0& 18.5&27.2 & 52.5 &18.4 &27.2 & 50.4 &16.7 & 25.0\\
5) P\&A&P& \cmark &\xmarkg&  51.9 &18.8 &27.6 & 52.3 & 18.6& 27.4 & 50.2 &16.0 & 24.2\\
6) \textbf{\PAD}& P &\cmark& \cmark & \textbf{52.1} & \textbf{19.6} & \textbf{28.4} & \textbf{52.6} &\textbf{19.2} & \textbf{28.1} & \textbf{50.5} &\textbf{18.5} & \textbf{27.0}\\
  \hline
  \end{tabular}
\vspace{-0.3cm}
\caption{Ablation study on ZSD, ZSI, and ZSS. G, P, A, D denote GMMN generator, primitive generator, disentanglement, and alignment, respectively. The results validate our goal of training a single model for universal zero-shot image segmentation tasks.}\label{tab:SOTA_2}
\vspace{-0.7cm}
\end{center}
\end{table*}
\subsection{Ablation Study}

In \cref{tab:SOTA_1} and \cref{tab:SOTA_2}, 
We perform ablation studies of the proposed \PAD on MS-COCO dataset under four tasks, including zero-shot panoptic segmentation, zero-shot instance segmentation, zero-shot object detection, and zero-shot semantic segmentation.
It is worth noting that the results in \cref{tab:SOTA_2} are obtained by the model trained on zero-shot panoptic segmentation task only, which achieves our goal of training a single model for universal zero-shot image segmentation tasks. For simplicity, our ablation analysis mainly focuses on ZSP, because ZSI, ZSD, ZSS have similar trends with ZSP. First, to demonstrate the advantage of introducing generative model, we implement a projection-based segmentation baseline by using CLIP text embeddings as classifier's weights, similar with ZegFormer-seg~\cite{ZegFormer}.
During training, there are 119 text embeddings used in classifier, while during inference, we add another 14 unseen text embeddings into classifier and label each object to one of these 133 classes.
As the 2nd row in \cref{tab:SOTA_1}, there is a strong bias towards seen classes, resulting in extreme low accuracy even zero for unseen group. 
Next, we construct baseline build upon generative GMMN model following ZS3~\cite{ZS3}, which outperforms projection-based method by 4.9\% in terms of unseen PQ. This phenomenon shows that generative model contributes to solving crucial bias issue.

\begin{figure}[t]
    \centering
	\includegraphics[width=0.49\textwidth]{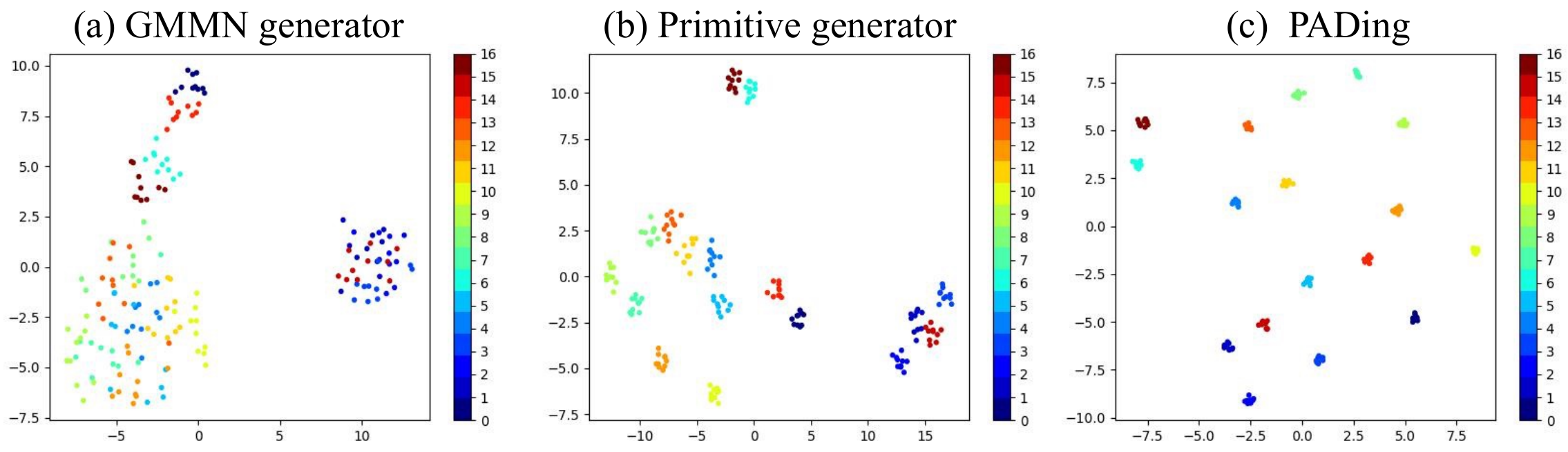}
	\vspace{-0.76cm}
	\caption{t-SNE visualization of synthesized features for unseen classes. (a) GMMN generator network.~(b) The proposed primitive generator.~(c) Our \PAD which further utilizes relationship information based on the primitive generator.}
	\vspace{-0.36cm}
	\label{fig:syn_feature}
\end{figure}

\textbf{GMMN \vs Our Primitive Generator.}
As shown in \cref{tab:SOTA_1} and \cref{tab:SOTA_2}, our primitive generator significantly surpasses GMMN generator by at least 9.0\% PQ, 5.7\% SQ, and 10.9\% RQ for HM metric. This shows that our primitive generator is capable of generating more effective features and the primitives can better grasp the real distribution of visual features compared to the baseline generator GMMN.
\vspace{-0.5cm}
\begin{table}[ht]
\setlength\tabcolsep{6pt}
\footnotesize
  \begin{center}
  \begin{tabular}{c|ccccccc}
  \hline
  \#Primitives & 100  & 200 & 300 & 400  & 500 & 600  & 700\\
  \hline
   PQ& 7.3 & 9.1& 10.2 &11.5 & 11.5 & 11.3 & 11.2\\
  \hline
  \end{tabular}
\vspace{-0.24cm}
\caption{\label{tab:number_ablation}Ablation study on the number of primitives.}
\vspace{-0.8cm}
\end{center}
\end{table}

\textbf{Number of Primitives.} We report the network's performance with different numbers of primitives in \cref{tab:number_ablation}. From the results, increasing the primitive number from 100 to 400 brings a significant performance gain of 4.2\%. The performance is a little down when the primitive number is larger than 400, thus we choose 400 as the default setting.

\textbf{Effectiveness of Alignment.}
Then, by applying semantic alignment as a constraint to our generator, the HM-PQ is further improved by 2.6\%, demonstrating the effectiveness of introducing inter-class relationships inherent from semantic space. 
Finally, we evaluate the alignment module with disentanglement, see 6) \PAD in \cref{tab:SOTA_1} and \cref{tab:SOTA_2}.
In comparison to using alignment only, alignment+disentanglement transfers semantic prior knowledge on semantic-related features and consistently brings performance gains of 2.0\% HM-PQ, 13.3\% HM-SQ, and 2.7\% HM-RQ.
The significant improvement demonstrates that the semantic-visual discrepancy has been alleviated owing to omitting semantic-unrelated noises. The utilization of disentanglement enables more effective alignment in the separated semantic-related space.

\textbf{Visualization of synthesized feature representations.}
To study the properties of our synthesized unseen features and demonstrate the effectiveness of our proposed approach, we employ t-SNE \cite{tsne} to show the distribution of our synthetic features in \cref{fig:syn_feature}. As we can see in \cref{fig:syn_feature}~\!(a), the synthesized features produced by GMMN generator are messy due to the semantic-visual discrepancy. In \cref{fig:syn_feature}~\!(b), when introducing our primitive generator, features belonging to the same class become more compact and features from different classes are highly separable. 
Furthermore, after applying relationship-alignment constraint on the semantic-related feature, see \cref{fig:syn_feature}~\!(c), features belonging to different classes are farther apart with better-structured distributions, which shows that the structure relationship is embedded into synthetic features and the synthesized unseen features are greatly enhanced with better discrimination.

\begin{figure*}[t]
    \centering
	\includegraphics[width=0.996\textwidth]{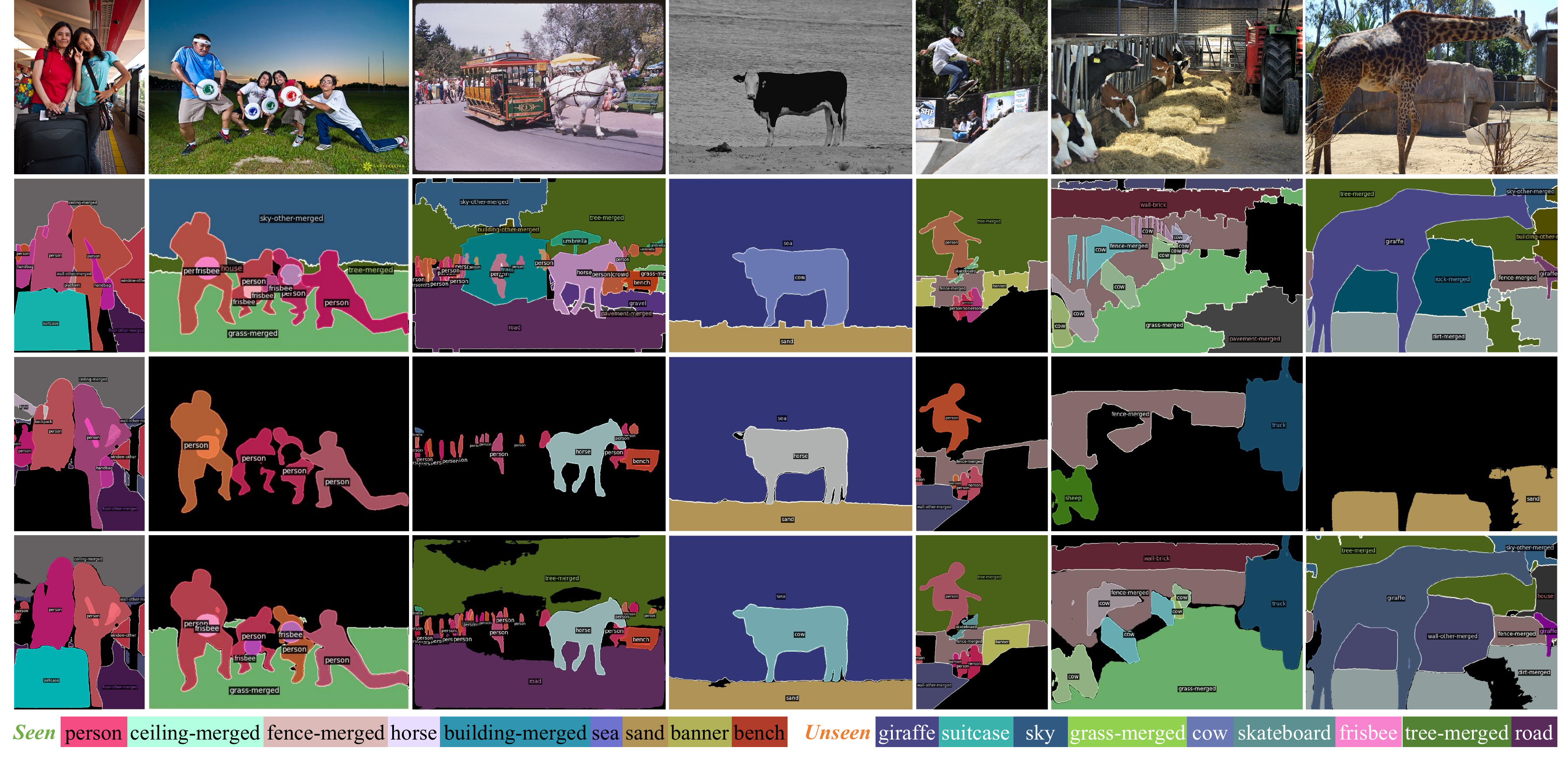}
	\vspace{-0.76cm}
	\caption{Qualitative results on the COCO for ZSP. The first row presents input images and the subsequent
rows illustrate ground-truth masks, predictions of the baseline, and predictions of our \PAD.} 
	\vspace{-0.36cm}
	\label{fig:visualization}
\end{figure*}

\vspace{-1mm}
\subsection{Comparison with State-of-the-art ZSS Methods}\vspace{-2mm}
To further validate the superiority of our approach, we compare it with previous state-of-the-art ZSS methods
on the challenging semantic segmentation datasets COCO-Stuff in \cref{tab:SOTA_ZSS}.
It is worth noting that we only report results without self-training and without complicated crop-mask image preprocess utilized for CLIP image encoder for a fair comparison. We train our model with semantic segmentation annotations.
The proposed approach outperforms the previous best method ZegFormer-seg~\cite{ZegFormer} by 3.5\% HM-IoU and 3.4\% unseen-IoU, demonstrating its effectiveness.
It is worth noting that the above methods use ResNet-101 while we only use ResNet-50.

\vspace{-1mm}\subsection{Comparison with State-of-the-art ZSI Methods}\vspace{-2mm}
We compare the proposed method with the previous state-of-the-art method ZSI~\cite{zsi} under the Generalized Zero-Shot Instance Segmentation (GZSI) setting in \cref{tab:SOTA_GZSI}. Our model is trained with instance segmentation annotations for a fair comparison. We achieve new state-of-the-art performance on both 48/17 split and 65/15 split. For example, we surpass ZSI by 7.20\% HM-mAP and 5.27\% HM-Recall on 48/17 split.
It is worth noting that ZSI~\cite{zsi} uses ResNet-101 while we use ResNet-50. 

\begin{table}[tbp]
\setlength\tabcolsep{4pt}
\footnotesize
  \begin{center}
  \begin{tabular}{lc|ccc}
  \hline
Method& Embed & Seen IoU & Unseen IoU & HM IoU  \\
  \hline
  \hline
 SPNet~\cite{SPNet} & Word2vec &  35.2 &\ \ 8.7 & 14.0 \\
 ZS3~\cite{ZS3}&Word2vec &     34.7 &\ \ 9.5& 15.0 \\
  CaGNet~\cite{CaGNet} & Word2vec &    33.5 & 12.2 &18.2 \\
  SIGN~\cite{SIGN} & Word2vec &    32.3 &15.5& 20.9 \\
   Zsseg-seg~\cite{zsseg_baseline} & CLIP &    38.7 &\ \ 4.9 &\ \ 8.7 \\
   ZegFormer-seg~\cite{ZegFormer} &CLIP &    37.4 & 21.4 & 27.2 \\
   \ours &CLIP & \textbf{40.4} & \textbf{24.8}& \textbf{30.7}\\
  \hline
  \end{tabular}
\vspace{-0.36cm}
\caption{\label{tab:SOTA_ZSS}Comparison with other ZSS methods on COCO-Stuff.}
\vspace{-0.76cm}
\end{center}
\end{table}

\vspace{-1mm}\subsection{Qualitative Results}\vspace{-2mm}
To qualitatively demonstrate the effectiveness of our proposed approach, we visualize some examples of zero-shot panoptic segmentation results in \cref{fig:visualization}. The second row is ground-truth mask while the third and fourth rows are predicted masks by baseline and our proposed approach, respectively. 
We observe that our \PAD successively finds several unseen classes, \eg, \texttt{suitcase}, \texttt{grass}, \texttt{frisbee}, \texttt{road}, \texttt{tree}, \texttt{skateboard}, that are missed or misclassified by the baseline model. Besides, thanks to the class-agnostic mask generation ability of Mask2Former~\cite{mask2former}, our results show high-quality masks.

\renewcommand{\multirowsetup}{\centering}
\begin{table}[t]
\setlength\tabcolsep{3.6pt}
\footnotesize
  \begin{center}
  \begin{tabular}{cl|cccccc}
  \hline
  &  &\multicolumn{2}{c}{Seen} & \multicolumn{2}{c}{Unseen} & \multicolumn{2}{c}{HM} \\
Split &   Method  &  mAP  & Recall & mAP  & Recall & mAP & Recall \\
  \hline
  \hline
   \multirow{2}{*}{48/17}& ZSI~\cite{zsi} &  43.0&  64.4  &\ \ 3.6 & 44.9 &\ \ 6.7 & 52.9 \\
   & \ours &\textbf{53.0}  &\textbf{75.1}  &\ \ \textbf{8.0}  &\textbf{47.5}  & \textbf{13.9} &\textbf{58.2} \\

  \hline
   \multirow{2}{*}{65/15}& ZSI~\cite{zsi} & 35.7 & 62.5 & 10.4 & 49.9 & 16.2 & 55.5  \\
   & \ours &\textbf{41.8}  &\textbf{73.2}  &\textbf{13.9}  &\textbf{51.3}  &\textbf{20.9} &\textbf{60.3} \\
  \hline
  \end{tabular}
\vspace{-0.24cm}
\caption{Results on GZSI using word2vec embedding.}\label{tab:SOTA_GZSI}
\vspace{-0.9cm}
\end{center}
\end{table}

\vspace{-2mm}\section{Conclusion}\vspace{-2mm}
We propose primitive generation with collaborative relationship alignment and feature disentanglement learning (\PAD) as a unified framework to achieve universal zero-shot segmentation. A primitive generator is proposed to synthesize fake training features for unseen classes. A collaborative feature disentanglement and relationship alignment learning strategy is proposed to help the generator produce better fake unseen features, where the former one decouples visual features to semantic-related part and semantic-unrelated part and the later one transfer inter-class knowledge from semantic space to visual space. Extensive experiments on three zero-shot segmentation tasks demonstrate the effectiveness of the proposed approach.

\footnotesize{\noindent\textbf{Acknowledgement} Shuting He and Wei Jiang were partially supported by National Natural Science Foundation of China (No.62173302).}

{\small
\bibliographystyle{ieee_fullname}
\bibliography{egbib}
}

\end{document}